\documentclass[letterpaper,journal]{IEEEtran}
\usepackage{amsmath,amsfonts}
\usepackage{algorithmic}
\usepackage{algorithm}
\usepackage{array}
\usepackage{textcomp}
\usepackage{stfloats}
\usepackage{url}
\usepackage{verbatim}
\usepackage{cite}
\usepackage{multirow}
\usepackage{latexsym}
\usepackage{color}
\usepackage{threeparttable}

\usepackage{multicol}      
\usepackage{booktabs}      
\usepackage{colortbl}      
\usepackage[table]{xcolor} 
\usepackage{graphicx}      
\usepackage{array}         
\usepackage{adjustbox}     
\usepackage{makecell}      
\usepackage{tabularx}      
\definecolor{Gray}{gray}{0.9}     
\usepackage{pifont}
\usepackage{calc}
\usepackage{threeparttable}

\usepackage{xcolor}
\usepackage{url}
\usepackage[colorlinks=false, pdfborder={0 0 0}]{hyperref}

\let\oldurl\url
\renewcommand{\url}[1]{\textcolor{green!50!black}{\oldurl{#1}}}

\begin{document}

\title{OZ-TAL: Online Zero-Shot Temporal Action Localization}

\author{Chaolei Han, Hongsong Wang, Xin Gong,~\IEEEmembership{Member,~IEEE}, and Jie Gui,~\IEEEmembership{Senior Member,~IEEE}
\thanks{This work was supported in part by the grant of the National Natural Science Foundation of China under Grant 62172090; Start-up Research Fund of Southeast University under Grant RF1028623097. (Corresponding author: Jie Gui and Hongsong Wang.)}
\thanks{Chaolei Han, Xin Gong, and Jie Gui are with the School of Cyber Science and Engineering, Southeast University, Nanjing 211102,
China. Jie Gui is also with Engineering Research Center of Blockchain Application, Supervision and Management (Southeast University), Ministry of Education; Purple Mountain Laboratories, Nanjing 211111, China (email: chaoleihan@seu.edu.cn, xingong@seu.edu.cn, guijie@seu.edu.cn).}
\thanks{Hongsong Wang is with the School of Computer Science and Engineering, Southeast University, Nanjing 211102, China (hongsongwang@seu.edu.cn).}}

\markboth{Journal of \LaTeX\ Class Files,~Vol.~14, No.~8, August~2021}%
{Shell \MakeLowercase{\textit{et al.}}: A Sample Article Using IEEEtran.cls for IEEE Journals}


\maketitle

\begin{abstract}
Online Temporal Action Localization (On-TAL) aims to detect the occurrence time and category of actions in untrimmed streaming videos immediately upon their completion. Recent advancements in this field focus on developing more sophisticated frameworks, shifting from Online Action Detection (OAD)-based aggregation paradigm to instance-level understanding. However, existing approaches are typically trained on specific domains and often exhibit limited generalization capabilities when applied to arbitrary videos, particularly in the presence of previously unseen actions.
In this paper, we introduce a new task called Online Zero-shot Temporal Action Localization (OZ-TAL), which aims to detect previously unseen actions in an online fashion. 
Furthermore, we propose a training-free framework that leverages off-the-shelf Vision-Language Models (VLMs) while introducing additional mechanisms to enhance visual representations and mitigate their inherent biases.
We establish new benchmarks and representative baselines for OZ-TAL on THUMOS14 and ActivityNet-1.3, and extensive experiments demonstrate that our method substantially outperforms existing state-of-the-art approaches under both offline and online zero-shot settings.
\end{abstract}


\section{Introduction}
\label{sec:intro}

\IEEEPARstart{B}{illions} of videos are generated and uploaded to online platforms and social media on a daily basis, thereby increasing the demand for advanced video understanding and analysis \cite{yu2024multi,yang2022video,li2024neighbor,luo2024adaptive,wu2024hypergraph,li2024efficient}. 
As a crucial technology, Online Temporal Action Localization (On-TAL) \cite{song2024online,reza2024hat,kang2024actionswitch} has garnered significant attention due to its promising applications across diverse scenarios, such as video surveillance, live sports broadcasting, and healthcare monitoring.

Unlike conventional temporal action localization (TAL)~\cite{zhao2022temporal,liu2022fineaction,tang2024learnable,zhao2025constructing,liu2025brtal}, On-TAL entails predicting both the start and end times of actions, as well as classifying their categories based solely on the input frames observed up to current frame \cite{wang2023temporal}. 
This task is particularly challenging due to the need for rapid predictions of actions that may occur simultaneously, with no possibility for retrospective modifications once action instances have been initially generated.

One primary approach for On-TAL is to aggregate the classification results of each frame generated by Online Action Detection (OAD) \cite{pang2025context,wang2024does}, which identifies ongoing actions in real-time video \cite{kang2024actionswitch,kim2022sliding}. 
Recent advancements shift from this paradigm to instance-level understanding, with a focus on memory mining through the development of more sophisticated transformer-based networks \cite{song2024online,reza2024hat}.
These architectures leverage long-term temporal relationship modeling to capture inter-frame dependencies, which helps distinguish activities that share similar appearances in certain frames.

\begin{figure}
  \centering
  \includegraphics[width=\linewidth]{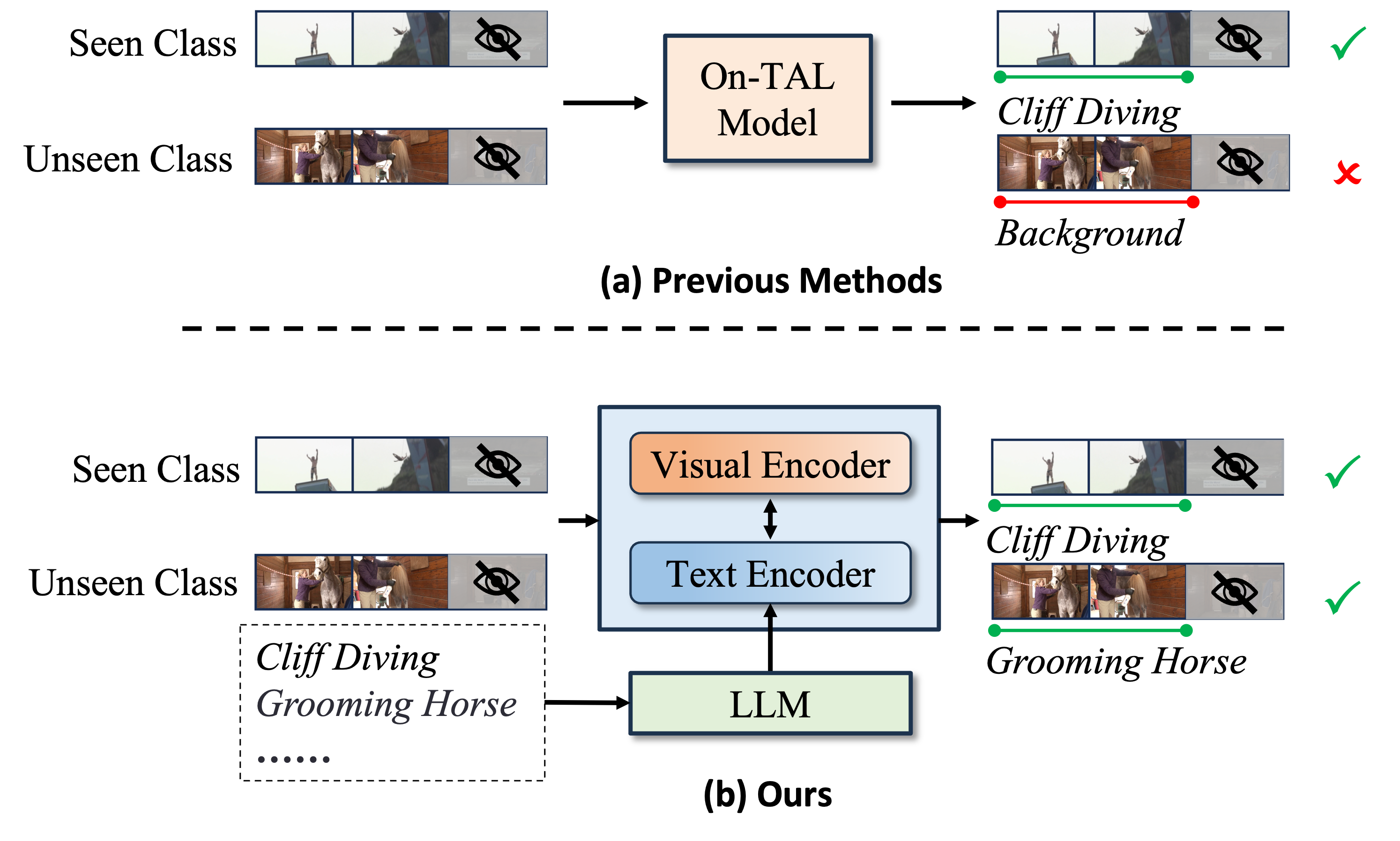}
  \caption{\textbf{Comparison between traditional On-TAL models and our method.}
    (a) Traditional models are constrained to recognizing only seen actions from training data, whereas (b) our approach leverages VLMs to detect arbitrary unseen actions with enhanced generalization.}
    \label{fig:difference}
\end{figure}

However, all existing approaches rely on frame-level annotations, which are both time-consuming and labor-intensive to obtain. Moreover, training on a specific dataset inevitably introduces domain bias, thereby hindering the model’s generalization to unseen data distributions, as illustrated in Figure~\ref{fig:difference}(a).
Consequently, we believe it is necessary to investigate the performance of On-TAL in open-set scenarios.
To this end, \textit{we introduce a novel task setting termed Online Zero-shot Temporal Action Localization (OZ-TAL), which aims to identify previously unseen actions in real time without requiring any task-specific training.} 

Vision-Language Models (VLMs) have gained significant prominence in zero-shot learning 
due to their strong generalization capability and extensive knowledge encapsulation. 
In the past, considerable efforts have been made to transfer these image-text pair pre-trained models to video understanding tasks.
In offline TAL, these models \cite{nag2022zero,ju2022prompting,raza2024zero} are typically employed as encoders to extract semantic information from raw videos and corresponding labels, followed by a learnable transformer to model long-term contextual dependencies. All previous work has inspired us to explore the application of VLMs’ strong generalization capabilities to OZ-TAL.
However, this task-specific training-free implementation faces two challenges:
(1) Although VLMs can perform frame-level classification on streaming videos, their limited capacity to model long-term dependencies hinders effective temporal context integration, leading to unreliable action localization.
\textcolor{black}{(2) In the training-free localization setting, VLMs are typically applied through direct frame-level or short-clip-level text-video matching. Such matching may be dominated by visually salient cues, such as human motion, pose changes, and object interactions, while scene-level and background semantics are insufficiently exploited. This issue becomes more pronounced when the discriminative evidence of an action is implicit or weakly motion-related.}


To address these limitations, as illustrated in Figure~\ref{fig:difference}(b), we propose a \textbf{V}LM-Based \textbf{F}eature-\textbf{E}nhanced \textbf{A}ction \textbf{L}ocalizer (VFEAL), which first performs frame-wise action classification and subsequently aggregates predictions into temporally localized action segments in real time.
Specifically, our action localizer first introduces a \textit{\textbf{memory-guided feature enhancement}} mechanism to strengthen the temporal interaction across frames. By dynamically incorporating memory of salient historical frames, VFEAL enriches the representation of current frame with informative visual cues, thereby mitigating noise distractions and enhancing temporal stability. 
To ensure computational efficiency, the fusion operates in a frame-to-segment manner, which avoids the linear growth in complexity with respect to the memory trace length.
To further address the limited background modeling capability of VLMs under unsupervised settings, VFEAL incorporates a \textit{\textbf{background-aware k-way classification}} strategy as a replacement for standard direct classification. 
This strategy leverages background semantics to assign stronger penalties to low-confidence predictions, while maintaining high confidence for clear-cut action categories. By suppressing ambiguous predictions and improving boundary sensitivity, this design effectively reduces the inherent bias of VLMs toward motion-centric features and enhances true negative performance.
Finally, VFEAL feeds the prediction results into an \textit{\textbf{online action span prediction}} to generate complete action instances. This module enforces temporal consistency and promotes the generation of high-quality action segments through a class-specific state machine.
We conduct extensive experiments on the THUMOS14 and ActivityNet-1.3 datasets to evaluate the effectiveness of VFEAL. Its simple yet effective design makes it well-suited for open-set inference over streaming videos. 

To summarize, our key contributions are as follows:

\begin{itemize}
\item \textbf{Novel task setting for overlapping action detection:} We propose a new task setting, termed OZ-TAL, which aims to detect the occurrence time and recognize the categories of overlapping actions in untrimmed streaming videos, strictly adhering to the constraints of no access to future information in open-world scenarios.
\item \textbf{Novel solution for OZ-TAL:} We introduce a VLM-based framework called VFEAL for the OZ-TAL task. It enhances frame-level features using historical context and refines classification with background semantics, enabling reliable action segment generation.
\item \textbf{New benchmarks and optimal performance:} We set up two new benchmarks for OZ-TAL on the THUMOS14 and ActivityNet-1.3 datasets and establish representative baselines for this task. Extensive experiments show that our method significantly beats state-of-the-art methods for both offline and online zero-shot TAL.

\end{itemize}

\section{Related Work}
\label{sec:relatedwork}

\noindent{\textbf{Online Temporal Action Localization:}}
Temporal Action Localization (TAL) \cite{zhao2022temporal,liu2022fineaction,tang2024learnable,chen2025temporal,XuSGLPG24,zeng2024benchmarking,zeng2024unimd,yang2024adapting,liu2024end,zhu2024dual,yang2024dyfadet} seeks to identify action segments within untrimmed videos and garners significant attention in the video understanding community. Building upon this foundational task, Online Temporal Action Localization (On-TAL) \cite{kang2024actionswitch,song2024online,kim20222pesnet} presents an even greater challenge, as it involves the inability to access future frames and the necessity of predicting action instances immediately upon their completion, without any retrospective modifications.

One straightforward approach is to adopt online action detection (OAD) \cite{pang2025context,wang2024does,cao2023e2e,an2023miniroad,yang2022colar}, which classifies frames individually in streaming videos, followed by aggregating the results over time.
Kang et al.~\cite{kang2021cag} first introduce On-TAL and extend OAD by employing a Markov Decision Process to integrate both frame features and historical decision context.
Tang et al.~\cite{tang2022simon} integrate both current and memory information within a transformer, enabling the detection of simultaneous actions in an end-to-end manner.
Kim et al.~\cite{kim2022sliding} propose an anchor-based method with a sliding window scheme, providing instance-level context for accurately grouping per-frame predictions.
Kang et al.~\cite{kang2024actionswitch} propose a class-agnostic framework by designing an ActionSwitch to record the state of action occurrences, enabling the detection of overlapping actions without relying on threshold-based grouping. 
Another effective method is to directly generate the action segment in a single step, using the current moment as the reference point.
Song et al.~\cite{song2024online} selectively preserve past information in a memory queue, which is scanned to identify the action start time, allowing the model to consider long-term context.
Reza et al.~\cite{reza2024hat} propose an action anticipation-guided history refinement method and a gradient-guided focal loss, which effectively enhance contextual relevance and mitigate class imbalance issues. 
Existing research has typically focused on training models for specific activities in a fully supervised manner, which struggles to generalize to arbitrary videos. In contrast to previous studies, we extend On-TAL to open scenarios, operating without any supervision or training.\\

\noindent{\textbf{Zero-Shot Temporal Action Detection:}}
Zero-shot Temporal Action Detection (ZS-TAD) \cite{nag2022zero,ju2022prompting} aims to identify action categories that have not been encountered during training, which requires the model to possess strong generalization capabilities across diverse actions. The core principle of zero-shot learning is to extract shared knowledge from prior information and transfer it from seen classes to unseen classes.
Zhang et al.~\cite{zhang2020zstad} are the first to achieve ZS-TAD by encoding seen and unseen activities using Word2Vec \cite{mikolov2013efficient}, effectively capturing shared semantic information. They further enhance label embeddings by incorporating the CLIP text encoder \cite{radford2021learning}, which leads to improved performance \cite{zhang2022tn}.
Nag et al.~\cite{nag2022zero} optimize classification and localization simultaneously with the aid of CLIP, effectively mitigating the error propagation problem and introducing a novel framework perspective for ZS-TAD.
Raza et al.~\cite{raza2024zero} develop mProTEA, leveraging layer-wise multimodal prompts and text-enhanced actionness priors to enable accurate and scalable zero-shot temporal action localization.
Liberatori et al.~\cite{liberatori2024test} adapt latent features extracted by CoCa \cite{yu2022coca} at test time, effectively addressing the issue of out-of-distribution data.
Inspired by these works, we introduce a novel task that aims to detect actions in real time under a zero-shot setting.\\

\noindent\textbf{Vision-Language Models for TAL:}
The success of large language models (LLMs) \cite{chowdhery2023palm,touvron2023llama} in the natural language processing (NLP) field catalyzes the development of multimodal large language models (MLLMs) in the computer vision domain \cite{zhu2023minigpt,achiam2023gpt,li2022blip,li2023blip,liu2024improved}. 
Subsequently, researchers have extended the visual modality from images to videos, developing MLLMs \cite{chen2024sharegpt4video,li2023videochat} specifically aimed at video understanding and analysis.
Notable works, such as ShareGPT4Video \cite{chen2024sharegpt4video} and VideoChat \cite{li2023videochat}, integrate advanced text and image understanding to facilitate rich dialogues, enabling these models to interpret visual content and generate detailed textual descriptions seamlessly.
Although these models excel in video question answering and reasoning tasks, their performance on TAL is limited, primarily due to the lack of segment constraints during training and their limited computational resources for processing long untrimmed videos effectively.

To address these issues, some works focus on designing architectures based on the backbone of VLMs.
For instance, 
Ju et al.~\cite{ju2022prompting} first adapt CLIP with a learnable prompt tailored for TAL, revealing the significant potential of VLMs in this task.
Li et al.~\cite{li2024detal} propose a decoupled network that performs class-agnostic detection using visual features extracted by CLIP, followed by a classification module that incorporates action-aware text features.
Lee et al.~\cite{lee2024text} leverage cross-attention to integrate visual and textual information by CLIP, followed by a foreground-aware head designed to emphasize discriminative sub-actions, 
resulting in a more effective network.
Building on prior work, we make the first attempt to harness the zero-shot capability of VLMs for action localization in streaming video, while strictly adhering to the constraint of avoiding retrospective modifications.

\section{Online Zero-Shot TAL}
\begin{figure}
  \centering
  \includegraphics[width=\linewidth]{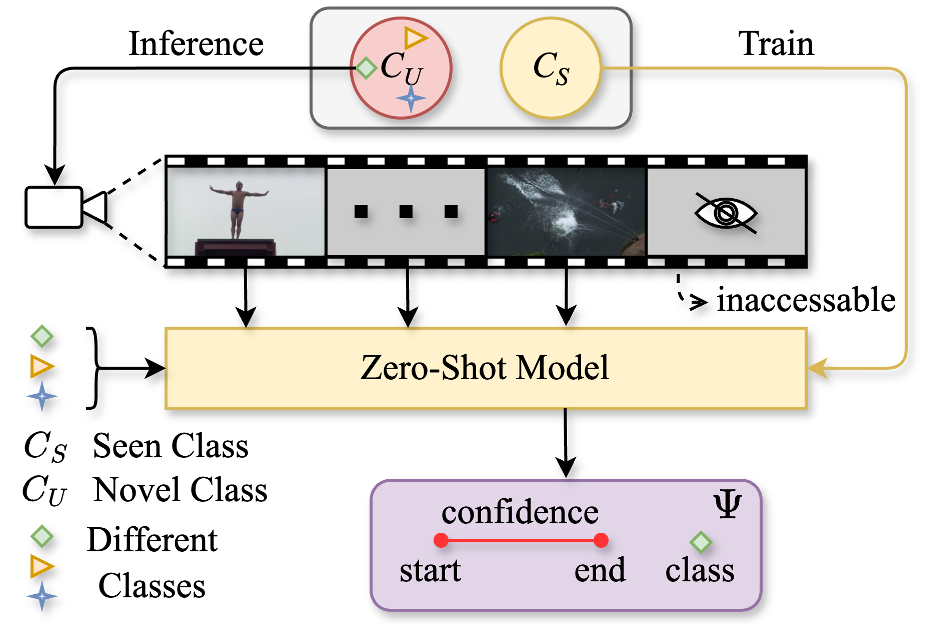}
  \caption{\textbf{Illustration of OZ-TAL.} The task involves predicting the start time, end time, and category of each action under two constraints: (1) the action categories in the training and test sets are completely disjoint, and (2) future frame information and post-processing are not available during inference.}
  \label{fig:oztal}
\end{figure}

In this section, we formally define the task of 
OZ-TAL,
as illustrated in Figure~\ref{fig:oztal}. We begin by reviewing the conventional TAL task and then introduce our proposed extension.
Given an untrimmed video sequence $\mathcal{V} = \left \{ v_i \right \}_{i=1}^{T}$ consisting of $T$ consecutive frames, the objective of conventional TAL is to predict a set of action instances $\mathit{\Psi}=\left \{ (s_n,e_n,c_n,p_n) \right \} _{n=1}^{N} $, where $s_n$ and $e_n$ denote the start and end timestamps of the $n$-th action, $c_n \in \mathcal{C}$ is the action category, and $p_n$ is the associated confidence score.
The proposed OZ-TAL task extends this formulation by introducing two constraints: zero-shot learning and online inference. Under the zero-shot setting, the action category set $\mathcal{C}$ is partitioned into two disjoint subsets: a set of seen classes $\mathcal{C}_s$ and a set of unseen classes $\mathcal{C}_u$, such that $\mathcal{C}_S \cap \mathcal{C}_U = \emptyset$. The model is trained exclusively on $\mathcal{C}_s$ and is expected to localize actions from $\mathcal{C}_u$ during inference.
Meanwhile, the online constraint requires the model to operate in a streaming setting, where only the current and past frames are accessible at each time step. Once an action instance is completed, the prediction must be generated immediately, with no opportunity for retrospective refinement.

These constraints make OZ-TAL a particularly challenging problem. The zero-shot condition demands strong generalization to previously unseen action categories, while the online requirement precludes access to future context, limiting the model’s ability to make globally informed decisions. As a result, OZ-TAL necessitates both robust temporal reasoning and effective utilization of prior knowledge to achieve accurate, real-time action localization. 
\textcolor{black}{In the next section, we propose VFEAL, which satisfies the OZ-TAL task setting without requiring task-specific training.}

\section{Method} \label{sec:method}

\begin{figure*}
  \centering
  \includegraphics[width=\linewidth]{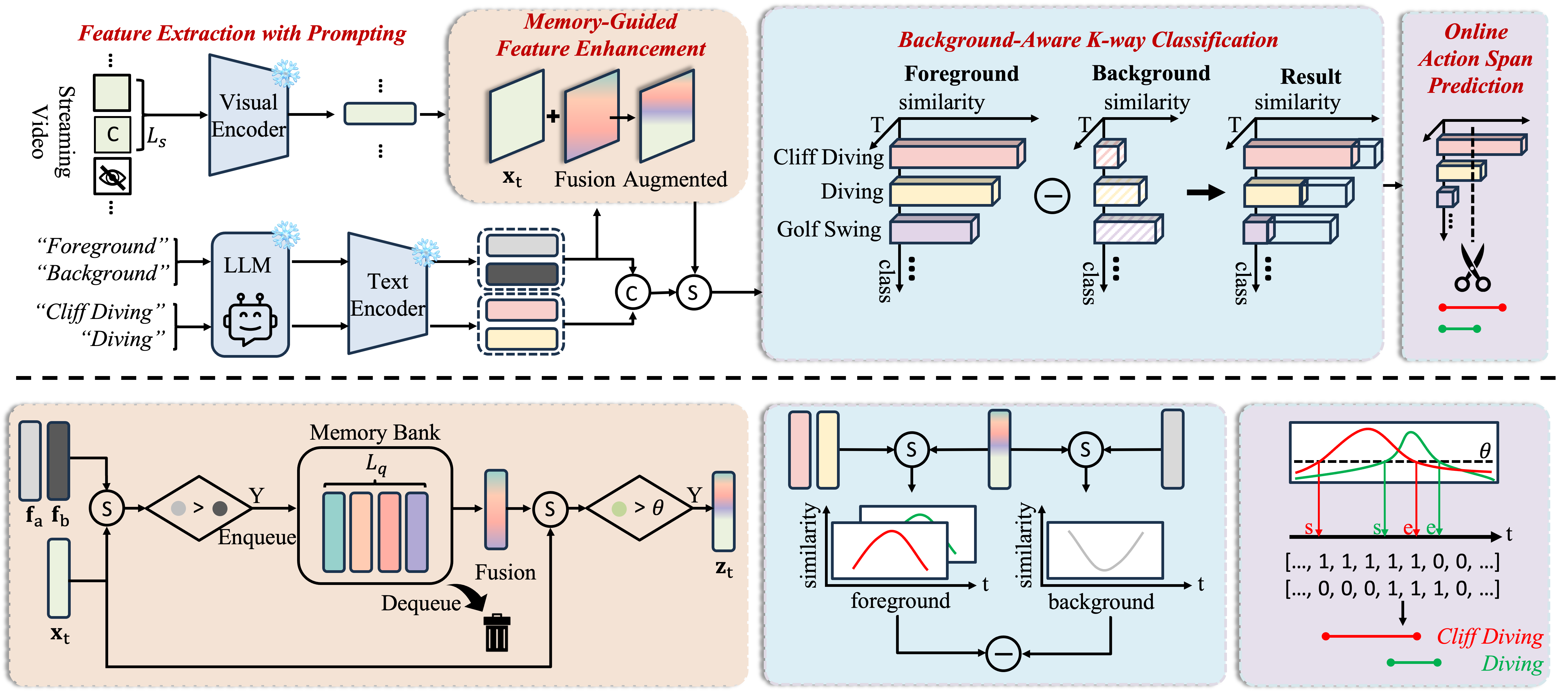}
  \caption{\textbf{Overview of the VLM-Based Feature-Enhanced Action Localizer (VFEAL).} The framework comprises four sequential components:
(1) Feature Extraction with Prompting: Short-term visual features and text features are extracted using the video and text encoders of a VLM;
(2) Memory-Guided Feature Enhancement: Long-term dependencies are modeled by enhancing current features with salient historical context;
(3) Background-Aware K-way Classification: Classification predictions are refined using background semantics to suppress ambiguous results;
(4) Online Action Span Prediction: Action segments with reliable confidence scores are generated based on frame-level classification outputs.
}
\label{fig:pipeline}
\end{figure*}

VFEAL is designed to address the OZ-TAL task by jointly performing frame-wise classification and instance-wise span prediction in an online manner. As illustrated in Figure~\ref{fig:pipeline}, the entire framework consists of four sequential components that function without the need for task-specific training.


\subsection{Feature Extraction with Prompting} \label{sec:extraction}
VFEAL builds upon a pre-trained VLM for feature extraction. 
Specifically, we select ViCLIP \cite{wang2023internvid} for its temporal-spatial modeling capabilities, as it integrates a visual encoder $E_{v}$ and a text encoder $E_{t}$ to enable effective multimodal representation learning. 
VFEAL employs a sliding window scheme that moves frame by frame to produce frame-level classification results from visual and textual embeddings, followed by online span prediction.\\

\noindent{\textbf{Visual Embedding:}}
At a given time $t$, the input video segment $\mathcal{V}_t=\left \{ v_i \right \} _{i=t-L_s+1}^{t} $ is processed by $E_{v}$ to obtain frame-level representations $\mathbf{x}_t \in \mathbb{R} ^{D} $, where $L_{s}$ denotes the length of the input segment and $D$ is the feature dimension.
Unlike previous works \cite{song2024online, reza2024hat} that require setting different optimal window sizes (\emph{i.e.}, input segment length) for different datasets, we use a fixed size across various datasets.\\

\noindent{\textbf{Textual Embedding:}}
In VLMs, the construction of prompts directly influences the alignment between visual and textual representations \cite{wasim2023vita}.
For TAL tasks, effective linguistic expressions require distinctive descriptions of activity characteristics, including temporal structure, spatial positioning, and key postures, enabling the model to precisely localize target actions \cite{xu2025information}.
Leveraging these insights, we employ an LLM to generate precise and informative descriptions for each action category, and the detailed prompt sent to the LLM is shown in Figure~\ref{fig:prompt}.
Subsequently, the text encoder $E_{t}$ takes class-specific descriptions as input and generates textual features $\mathbf{F}_{cls} \in \mathbb{R} ^{K\times D}$, where $K$ denotes the number of categories in $\mathcal{C}_S$ or $\mathcal{C}_U$. 
\textcolor{black}{
In addition, two coarse semantic descriptions are constructed to represent foreground and background semantics. 
Here, foreground denotes the presence of action-related content, whereas background denotes frames or clips without target-action semantics. 
These prompts are not limited to sports scenes and can be instantiated according to the target domain. 
For example, in sports videos, they can be written as ``A scene depicting a player engaging in some sports activities'' and ``A scene without sports activities.'' 
These two descriptions are encoded as $\mathbf{f}_a \in \mathbb{R}^D$ and $\mathbf{f}_b \in \mathbb{R}^D$, respectively, to represent foreground and background semantics.
}

\subsection{Memory-Guided Feature Enhancement} \label{sec:fusion}
Although the visual feature of each frame is encoded with short-term content (\emph{i.e.}, $L_s$ consecutive frames), they lack equally crucial long-term historical context.
The goal of Memory-Guided Feature Enhancement (MGFE) is to selectively retain essential historical information to enhance the current visual feature. At each time step, the current visual feature is processed through two components of this fusion module: the memory bank and memory enhancement.\\

\noindent{\textbf{Long-Term Memory Bank:}} In online streaming video processing, efficiently storing and rapidly retrieving past frame information is crucial for accurately detecting action instances, particularly important for identifying long-duration actions that extend beyond the current input segment \cite{song2024online,reza2024hat}.
Due to the disparity in timescales between input segment length and action duration, accurately identifying long-duration actions requires cross-temporal association analysis over multiple segments. In contrast, single-segment detection methods struggle to capture the complete temporal structure of an action, leading to incomplete or fragmented recognition.
To address this issue, we employ a memory queue to construct a long-term memory bank that selectively preserves valuable historical information under the specified conditions.
Specifically, a class-agnostic binary classification is performed on $\mathbf{x}_t$ to coarsely determine whether the current frame belongs to the foreground or background.
Given a memory bank $\mathcal{Q} = [\mathbf{q}_1,\mathbf{q}_2,\dots,\mathbf{q}_m]$ with a maximum capacity of $L_q$, the update process can be formulated as:
\begin{equation}
\mathcal{Q}=
\begin{cases}
    [\mathbf{q}_1,\mathbf{q}_2,\dots,\mathbf{q}_m,\mathbf{x}_t], & \text{if }\cos(\mathbf{x}_t,\mathbf{f}_{b}) < \cos(\mathbf{x}_t,\mathbf{f}_{a}); \\
    \makebox[\widthof{$[\mathbf{q}_1,\mathbf{q}_2,\dots,\mathbf{q}_m,\mathbf{x}_t]$}][c]{\phantom{-}$\mathcal{Q}$}, & \text{otherwise,}
\end{cases}
\end{equation}
where $\cos(\cdot,\cdot)$ denotes the cosine similarity. $\mathbf{f}_{b}$ and $\mathbf{f}_{a}$ represent the semantics for background and foreground, respectively. If a memory overflow occurs, the earliest stored elements in the memory queue will be discarded.\\

\noindent{\textbf{Feature Enhancement:} }
In an online setting, streaming videos may be affected by noise factors such as occlusion and camera shake, leading to instability in single-frame classification and impacting the subsequent segmentation process. Leveraging long-term memory can effectively mitigate these effects by providing temporal smoothing.
Most approaches leverage cross-attention to capture valuable contextual information, effectively tracking the temporal span of actions \cite{tang2022simon,reza2024hat}. 
However, this frame-to-frame attention mechanism introduces computational complexity that scales linearly with the length of the memory trace, 
which poses challenges for real-time inference.
To this end, we introduce a frame-to-segment fusion method that incurs negligible additional computational complexity.
Specifically, the updated memory bank $\mathcal{Q}$ is first averaged to generate a representative feature as:
\begin{equation}
    \mathbf{\bar{q}}_t=\frac{1}{L_q} \sum_{i=1}^{L_q} \mathbf{q}_i , \quad \mathbf{q}_i\in \mathcal{Q}.
    \label{eq2}
\end{equation}
If the similarity score between $\mathbf{\bar{q}_t}$ and $\mathbf{x}_t$ exceeds the threshold $\theta$, it indicates that the memory bank contains rich visual information relevant to the current frame action, which can be leveraged to enhance the robustness of the current feature.
Subsequently, we employ a similarity-based dynamic fusion mechanism that integrates information from both long-term memory and short-term features. 
Given that frames from different time steps contribute unequally to current predictions \cite{gritsenko2024end}, the visual features in the memory bank are reweighted according to their temporal proximity to the current frame, with closer frames assigned higher weights, as formulated by:
\begin{equation}
    \mathbf{\tilde{q}}_t=\frac{1}{L_q} \sum_{i=1}^{L_q} \frac{1}{L_q+1-i} \mathbf{q}_i , \quad \mathbf{q}_i\in\mathcal{Q}.
    \label{eq3}
\end{equation}
\textcolor{black}{
The standard mean in Eq.~\ref{eq2} provides a stable global summary for memory relevance estimation, whereas the temporally weighted mean in Eq.~\ref{eq3} emphasizes recent memory features for feature fusion.
}
The reweighted features are then aggregated to generate a segment-level representation, which is utilized to enhance the current feature. This process can be formulated as:
\begin{equation}
    \lambda_t = \frac{1}{2} \mathrm{norm}(\cos(\mathbf{x}_t,\mathbf{\bar{q}}_t)),
\end{equation}
\begin{equation}
    \mathbf{z}_t =(1-\lambda_t)\mathbf{x}_t+\lambda_t \mathbf{\tilde{q}_t},
\end{equation}
where norm denotes normlization, $\mathbf{\tilde{q}_t}\in\mathbb{R}^{D}$  is the segment-level representation, and $\lambda_t$ is the co-efficient used to adaptively regulate the degree of fusion. 
Notably, if the similarity value is below the threshold, $\mathbf{z}_t=\mathbf{x}_t$.
The fused feature $\mathbf{z}_t\in \mathbb{R}^{D}$ is then forwarded to the subsequent stage.

\subsection{Background-Aware K-way Classification}\label{subsec:classification}
Although action probabilities can be derived by measuring the cosine similarity between class-specific action queries and visual features, directly aggregating these $K$-way classification results leads to unsatisfactory performance.
In fully supervised settings, a randomly initialized background query is typically introduced alongside class-specific action queries, transforming the task into a $(K+1)$-way classification problem \cite{nag2022zero,nag2023semantics}. However, this approach is infeasible for our training-free method, which operates without annotations.
Moreover, VLMs inherently exhibit learning bias caused by the data distribution encountered during pre-training.
Without fine-tuning on the target domain, the model prioritizes action category recognition over background differentiation. For instance, scene transition features, once captured, are often misinterpreted as motion-related cues.
Therefore, we propose constraining the $K$-way classification results using background confidence instead of introducing an additional background category.
To preserve significant scores that are often associated with the true class while suppressing irrelevant ones, we adaptively incorporate background awareness to generate the final classification results. Mathematically, this process can be written as follows:
\begin{equation}
    \alpha_t = \max(\frac{\mathbf{k}_t}{\mathbf{k}_t+r_t \mathbf{I}}, 0.5),
\end{equation}
\begin{equation}
    \mathbf{y}_t = \alpha_t \mathbf{k}_t - (1-\alpha_t) r_t \mathbf{I},
\end{equation}
where $t$ denotes the time step and $\mathbf{I} \in \mathbb{R}^{K}$ is an all-ones vector. 
\textcolor{black}{Here, $\mathbf{k}_t \in \mathbb{R}^{K}$ denotes the $K$-dimensional action classification score vector at time step $t$, obtained by computing the cosine similarity between the enhanced visual feature $\mathbf{z}_t$ and the class-specific textual features $\mathbf{F}_{cls}$.}
The background confidence $r_t \in \mathbb{R}$ is computed as the cosine similarity between $\mathbf{z}_t$ and $\mathbf{f}_b$.
Once the final classification scores $\mathbf{y}_t \in \mathbb{R}^K$ are generated, they are sent to the following action span prediction.
\textcolor{black}{Note that $\mathbf{k}_t$ and $\mathbf{y}_t$ denote scaled matching logits rather than raw cosine similarities, and therefore their values are not restricted to $[-1,1]$.}

\begin{table*}[t]
\caption{
Results of OZ-TAL models on THUMOS14 at different tIoUs. ($\ast$) indicates methods originally designed for closed-set scenarios but adapted for open-set evaluation in this experiment. ($\dagger$) denotes results taken from \cite{liberatori2024test}. “OF” refers to optimization-free methods, while “OOD” denotes out-of-distribution.
}
\begin{adjustbox}{max width=\textwidth}
\centering
\begin{threeparttable}
\begin{tabular}{ccccccccccccccccc}
\toprule[1pt]
\multirow{2}{*}{\textbf{Setting}} & \multirow{2}{*}{\textbf{Methods}} & \multirow{2}{*}{\shortstack{\textbf{Visual} \\ \textbf{Encoder}}} & \multirow{2}{*}{\textbf{OF}} & \multirow{2}{*}{\textbf{OOD}} & \multicolumn{6}{c}{\textbf{75\%-25\%}}                                                          & \multicolumn{6}{c}{\textbf{50\%-50\%}}                                     \\ \cmidrule(lr){6-11} \cmidrule(lr){12-17}
                         &                          &                                 &                 &            & 0.3   & 0.4   & 0.5  & 0.6  & \multicolumn{1}{c|}{0.7}  & \multicolumn{1}{c|}{Avg} & 0.3   & 0.4   & 0.5  & 0.6  & \multicolumn{1}{c|}{0.7}  & Avg \\ \midrule
\multirow{4}{*}{offline} & \multicolumn{1}{l|}{$\mathrm{EffPrompt^\dagger}$ \cite{ju2022prompting}}                & \multicolumn{1}{c|}{CLIP}                          & \multicolumn{1}{c|}{\ding{55}}  & \multicolumn{1}{c|}{\ding{51}}    & 7.1   & 5.9   & 4.5  & 3.4  & \multicolumn{1}{c|}{2.2}  & \multicolumn{1}{c|}{4.6}     & 5.4   & 4.4   & 3.5  & 2.7  & \multicolumn{1}{c|}{1.9}  & 3.6     \\
                         & \multicolumn{1}{l|}{$\mathrm{STALE^\dagger}$ \cite{nag2022zero}}                    & \multicolumn{1}{c|}{CLIP}                          & \multicolumn{1}{c|}{\ding{55}}    & \multicolumn{1}{c|}{\ding{51}}  & 0.5   & 0.3   & 0.2  & 0.2  & \multicolumn{1}{c|}{0.2}  & \multicolumn{1}{c|}{0.3}     & 1.3   & 0.7   & 0.6  & 0.6  & \multicolumn{1}{c|}{0.4}  & 0.7     \\
                         & \multicolumn{1}{l|}{$\mathrm{T3AL_{T=0}}$ \cite{liberatori2024test}}                    & \multicolumn{1}{c|}{CoCa}                          & \multicolumn{1}{c|}{\ding{51}}  & \multicolumn{1}{c|}{\ding{51}}      & 11.1  & 6.5   & 3.2  & 1.5  & \multicolumn{1}{c|}{0.6}  & \multicolumn{1}{c|}{4.6}     & 11.4  & 6.8   & 3.5  & 1.7  & \multicolumn{1}{c|}{0.6}  & 4.8     \\
                         & \multicolumn{1}{l|}{T3AL \cite{liberatori2024test}}                     & \multicolumn{1}{c|}{CoCa}                          & \multicolumn{1}{c|}{\ding{55}}  & \multicolumn{1}{c|}{\ding{51}}      & 19.2  & 12.7  & 7.4  & 4.4  & \multicolumn{1}{c|}{2.2}  & \multicolumn{1}{c|}{9.2}     & 20.7  & 14.3  & 8.9  & 5.3  & \multicolumn{1}{c|}{2.7}  & 10.4    \\ \midrule
\multirow{7}{*}{online}  & \multicolumn{1}{l|}{$\mathrm{MATR^\ast}$ \cite{song2024online}}                    & \multicolumn{1}{c|}{TSN}                          & \multicolumn{1}{c|}{\ding{55}}  & \multicolumn{1}{c|}{\ding{51}}      & 0.86  & 0.79  & 0.76 & 0.52 & \multicolumn{1}{c|}{0.33} & \multicolumn{1}{c|}{0.65}    & 0.75  & 0.66  & 0.56 & 0.45 & \multicolumn{1}{c|}{0.26} & 0.54    \\
                         & \multicolumn{1}{l|}{$\mathrm{HAT^\ast}$ \cite{reza2024hat}}                   & \multicolumn{1}{c|}{TSN}                          & \multicolumn{1}{c|}{\ding{55}}  & \multicolumn{1}{c|}{\ding{51}}     & 0.78  & 0.75  & 0.62 & 0.53 & \multicolumn{1}{c|}{0.31} & \multicolumn{1}{c|}{0.60}    & 0.56  & 0.53  & 0.49 & 0.41 & \multicolumn{1}{c|}{0.32} & 0.46    \\
                         & \multicolumn{1}{l|}{$\mathrm{MATR^\ast}$ \cite{song2024online}}                    & \multicolumn{1}{c|}{ViCLIP}                          & \multicolumn{1}{c|}{\ding{55}}    & \multicolumn{1}{c|}{\ding{51}}   & 0.38  & 0.34  & 0.28 & 0.15 & \multicolumn{1}{c|}{0.10} & \multicolumn{1}{c|}{0.25}    & 0.38  & 0.31  & 0.19 & 0.14 & \multicolumn{1}{c|}{0.05} & 0.21    \\
                         & \multicolumn{1}{l|}{$\mathrm{HAT^\ast}$ \cite{reza2024hat}}                     & \multicolumn{1}{c|}{ViCLIP}                          & \multicolumn{1}{c|}{\ding{55}}  & \multicolumn{1}{c|}{\ding{51}}     & 0.33  & 0.28  & 0.18 & 0.09 & \multicolumn{1}{c|}{0.04} & \multicolumn{1}{c|}{0.18}    & 0.27  & 0.22  & 0.15 & 0.11 & \multicolumn{1}{c|}{0.05} & 0.16    \\
                         & \multicolumn{1}{l|}{Baseline-I}               & \multicolumn{1}{c|}{CLIP}                          & \multicolumn{1}{c|}{\ding{51}}  & \multicolumn{1}{c|}{\ding{51}}     & 5.88  & 3.3   & 1.48 & 1.06 & \multicolumn{1}{c|}{0.56} & \multicolumn{1}{c|}{2.46}    & 6.21  & 3.44  & 1.81 & 1.03 & \multicolumn{1}{c|}{0.52} & 2.60    \\
                         & \multicolumn{1}{l|}{Baseline-II}              & \multicolumn{1}{c|}{ViCLIP}                          & \multicolumn{1}{c|}{\ding{51}}  & \multicolumn{1}{c|}{\ding{51}}     & 6.54  & 4.16  & 2.96 & 2.26 & \multicolumn{1}{c|}{1.54} & \multicolumn{1}{c|}{3.49}    & 6.19  & 4.01  & 2.64 & 2.06 & \multicolumn{1}{c|}{1.23} & 3.23    \\
                         & \multicolumn{1}{l|}{VFEAL (ours)}                    & \multicolumn{1}{c|}{ViCLIP}                          & \multicolumn{1}{c|}{\ding{51}}  & \multicolumn{1}{c|}{\ding{51}}     & 20.11 & 12.75 & 7.61 & 4.09 & \multicolumn{1}{c|}{1.96} & \multicolumn{1}{c|}{9.30}    & 19.51 & 12.62 & 7.71 & 4.31 & \multicolumn{1}{c|}{2.05} & 9.24    \\ 
\bottomrule[1pt]
\end{tabular}
\end{threeparttable}
\end{adjustbox}
\label{table1}
\end{table*}

\subsection{Online Action Span Prediction} \label{sec:segmentation}
Kang et al. \cite{kang2024actionswitch} propose a class-agnostic ActionSwitch to detect overlapping actions. Inspired by this, we employ a class-specific state machine to generate final action segments and their corresponding confidence scores based on frame-wise classification results. 
As illustrated in Figure~\ref{fig:pipeline}, this state machine is represented by a binary state matrix $\mathbf{M} \in \mathbb{R}^{T \times K}$, where the classification results and the threshold jointly determine the machine’s state. Specifically, this state indicator $\mathbf{M}$ is initialized as an all-zero matrix. Once the final classification result at time step $t$ is obtained, it is immediately compared with the action threshold. If the result for a certain action exceeds the threshold, the corresponding action state transitions from 0 to 1, which can be represented as follows:
\begin{equation}
\mathbf{M}_{t,k}=
\begin{cases}
1,   & \text{ if } \mathbf{y}_{t,k} > \tau;\\
0, & \text{ otherwise } ,
\end{cases}
\end{equation}
where $\tau$ is action threshold. Once the state transitions from 1 to 0, it signifies the completion of an action. The final action instance $\mathit{\Psi}=\left \{ (s,e,c,p) \right \} $ is generated immediately, where confidence score $p$ is normalized as in Eq.~\ref{eq9}. 
\begin{equation}
    p=\frac{\sum_{i=s}^{e} \mathbf{y}_{i,k}}{\sqrt{e-s+1} }.
\label{eq9}
\end{equation}
To prevent long intervals from inflating actionness simply by accumulating many high-CLIP frames, we apply sublinear scaling so the marginal contribution of each additional similarity score diminishes as counts grow, placing greater emphasis on the aggregate evidence rather than duration.


\section{Experiments}
\subsection{Experimental Setting}

\noindent{\textbf{Dataset:}} 
Experiments are conducted on two widely used datasets for TAL. Specifically, THUMOS14 \cite{idrees2017thumos} comprises 200 training videos and 213 test videos, covering 20 sports activities. 
On average, each video contains 15.5 action segments.
ActivityNet-1.3 \cite{caba2015activitynet} includes 200 daily activities with a total of 19,994 videos.
On average, each video contains 1.5 action segments.
For the zero-shot setting, we follow the protocol established in previous works \cite{nag2022zero,liberatori2024test}, assigning 75\% of action classes for training and 25\% for testing, as well as 50\% of action classes for training and 50\% for testing.
All experimental results are averaged over 10 random splits.\\


\noindent{\textbf{Evaluation Metric:}}
Mean Average Precision (mAP) is a crucial metric for TAL tasks, as it evaluates both the temporal localization accuracy and the classification performance of detected actions. 
We compute mAP at multiple temporal Intersection over Union (tIoU) thresholds. For each threshold, mAP is obtained by averaging the AP across all classes. 
\textcolor{black}{
THUMOS14 is evaluated at tIoU thresholds $\{0.3, 0.4, 0.5, 0.6, 0.7\}$, while ActivityNet-1.3 is evaluated at tIoU thresholds $\{0.5, 0.75, 0.95\}$.
}\\

\noindent{\textbf{Implementation Details:}}
We select ViCLIP \cite{wang2023internvid} and DeepSeek-R1 \cite{guo2025deepseek} as the VLM and LLM, respectively. Each video frame is resized to a resolution of $224 \times 224$ before being fed into the model.
We set the input sequence length $L_s$ and the fusion threshold $\theta$ to 8 and 0.8, respectively, for both datasets. 
The memory bank length $L_q$ is set to 20 and 40 for THUMOS14 and ActivityNet-1.3, respectively, and the classification threshold $\tau$ is set to 10 and 8 accordingly.
The entire framework is implemented in PyTorch and runs on an NVIDIA RTX 4090 GPU.

\begin{table}[t]
\caption{
Results of OZ-TAL models on ActivityNet-1.3. 
}
\begin{adjustbox}{max width=\columnwidth}
\centering
\begin{tabular}{clc|ccc|c}
\toprule[1pt]
\textbf{Split}                                                                                & \multicolumn{1}{c}{\textbf{Methods}} & \textbf{Visual Encoder} & \textbf{0.5} & \textbf{0.75} & \textbf{0.95} & \textbf{Avg} \\ \midrule
\multicolumn{1}{c|}{\multirow{3}{*}{\begin{tabular}[c]{@{}c@{}}75\%\\ |\\ 25\%\end{tabular}}} & Baseline-I                           & CLIP                    & 0.89         & 0.28          & 0.09          & 0.42             \\
\multicolumn{1}{c|}{}                                                                         & Baseline-II                          & ViCLIP                  & 1.16         & 0.35          & 0.12          & 0.54             \\
\multicolumn{1}{c|}{}                                                                         & VFEAL                                & ViCLIP                  & 11.63        & 3.4           & 0.34          & 5.13             \\ \midrule
\multicolumn{1}{c|}{\multirow{3}{*}{\begin{tabular}[c]{@{}c@{}}50\%\\ |\\ 50\%\end{tabular}}} & Baseline-I                           & CLIP                    & 0.78         & 0.27          & 0.07          & 0.37             \\
\multicolumn{1}{c|}{}                                                                         & Baseline-II                          & ViCLIP                  & 1.13         & 0.33          & 0.11          & 0.52             \\
\multicolumn{1}{c|}{}                                                                         & VFEAL                                & ViCLIP                  & 11.28        & 3.38          & 0.31          & 4.99             \\ 
\bottomrule[1pt]
\end{tabular}
\end{adjustbox}
\label{table2}
\end{table}

\subsection{Main Results}
As pioneers in evaluating OZ-TAL, we propose two baselines adapted from pre-trained VLMs for comparison. 
Specifically, Baseline-I and Baseline-II adopt CLIP (ViT-B/16) \cite{radford2021learning} and ViCLIP (ViT-L/16) \cite{wang2023internvid} as their respective backbones. 
Both naive approaches compute the cosine similarity between frame-level visual features and text representations of each action, followed by applying softmax to obtain classification scores and aggregating them across frames. A threshold of 0.8 is applied to differentiate foreground from background.

\noindent{\textbf{Results on THUMOS14:}}
We extend two advanced closed-set On-TAL approaches \cite{song2024online,reza2024hat} to open scenarios by modifying dataset splits and classification head to align with the zero-shot setting. Specifically, we replace their original prediction heads with a VLM-based semantic matching mechanism using the same prompts and text encoder as our model. For fair comparison, we evaluate both TSN and ViCLIP as visual encoders and keep all VLM and LLM configurations consistent.
Table~\ref{table1} compares the performance of our VFEAL model against two baselines and several state-of-the-art methods.
We also include several offline results for reference.
Although the online setting is more stringent than the offline one, our method achieves comparable performance to T3AL (9.3 \emph{vs.} 9.2 and 9.24 \emph{vs.} 10.4), which is supervised by pseudo-labels generated based on global features.
It can be observed that On-TAL methods designed for closed-set scenarios fail to transfer knowledge learned during training to novel actions, owing to inherent limitations in their network architectures.
While Baseline-I and Baseline-II outperform these methods, their overall performance remains suboptimal, suggesting that simple extensions or adaptations are inadequate to effectively address the challenges of OZ-TAL.
The proposed VFEAL consistently outperforms all baselines across both splits, demonstrating its superiority in the OZ-TAL setting.\\

\noindent{\textbf{Results on ActivityNet-1.3:}}
Since no existing On-TAL codebase is available for ActivityNet-1.3, we only compare OZ-TAL with the two baselines, as shown in Table \ref{table2}. 
Although the action distribution within each video in ActivityNet-1.3 is relatively simple, the high diversity across 200 action categories, coupled with the inability to leverage global prior knowledge of the video, poses a challenge to the paradigm of aggregating OAD results.
Under the same backbone, VFEAL surpasses Baseline-II by by over 4\%, demonstrating its ability to effectively leverage the zero-shot capabilities of off-the-shelf VLMs for streaming video.

\begin{table}[t]
\caption{Analysis of VFEAL components.}
\begin{adjustbox}{max width=\columnwidth}
\centering
\begin{tabular}{cccccccccc}
\toprule[1pt]
\textbf{Row}           & \textbf{Method}                             & \textbf{MGFE} & \textbf{BAKC} & \textbf{0.3} & \textbf{0.4} & \textbf{0.5} & \textbf{0.6} & \textbf{0.7} & \textbf{Avg} \\ \midrule
\multicolumn{1}{c|}{1} & \multicolumn{1}{l|}{Baseline-II}               & \ding{55}                     & \multicolumn{1}{c|}{\ding{55}}  & 6.6  & 3.7 & 1.5 & 2.0 & \multicolumn{1}{c|}{1.0} & 2.8    \\ \midrule
\multicolumn{1}{c|}{2} & \multicolumn{1}{c|}{\multirow{4}{*}{VFEAL}} &  \ding{55}                    & \multicolumn{1}{c|}{\ding{55}}  & 9.3  & 5.9 & 3.3 & 1.9 & \multicolumn{1}{c|}{1.0} & 4.3    \\
\multicolumn{1}{c|}{3} & \multicolumn{1}{l|}{}                       & \ding{51}                    & \multicolumn{1}{c|}{\ding{55}}  & 13.2 & 8.7 & 5.3 & 2.8 & \multicolumn{1}{c|}{1.4} & 6.3 \\
\multicolumn{1}{c|}{4} & \multicolumn{1}{l|}{}                       &   \ding{55}                   & \multicolumn{1}{c|}{\ding{51}} & 14.1 & 8.4 & 5.3 & 3.1 & \multicolumn{1}{c|}{1.6} & 6.5 \\
\multicolumn{1}{c|}{5} & \multicolumn{1}{l|}{}                       & \ding{51}                    & \multicolumn{1}{c|}{\ding{51}} & 17.7 & 11.6 & 7.0 & 3.9 & \multicolumn{1}{c|}{1.9} & 8.4    \\ 
\bottomrule[1pt]
\end{tabular}
\end{adjustbox}
\label{table3}
\end{table}

\subsection{Ablations and Analysis}
All experiments in this section, unless otherwise stated, are conducted on THUMOS14 using all 413 videos.\\

\noindent{\textbf{Component Analysis of VFEAL:}}
We perform ablation studies on the MGFE and BAKC modules to comprehensively analyze their structural contributions, as summarized in Table~\ref{table3}.
The baseline applies softmax normalization to per-frame classification results, followed by aggregating consecutive frames with scores exceeding 0.8.
Even without any online processing enhancements, our VFEAL outperforms Baseline-II by 1.5\% owing to its capability to detect simultaneously occurring actions.
Row 3 assesses the impact of using long-term memory in isolation to enhance short-term features, leading to a 2\% increase in mAP.
Row 4 demonstrates the effectiveness of suppressing classifications using background scores, leading to an additional 2.2\% performance gain.
When equipped with both components, VFEAL achieves the best performance across all evaluated videos under all tIoU thresholds.\\

\begin{table}[t]
\caption{Analysis of real-time performance. Experiments are conducted on untrimmed videos with an average duration of 212.68 s and a frame rate of approximately 30 FPS.}
\centering           
\begin{adjustbox}{max width=\columnwidth}
\begin{tabular}{ccccc}
\toprule[1pt]
\textbf{Row} & \textbf{Input} & \textbf{Visual Encoder} &\textbf{Inference Speed (FPS)} & \textbf{Latency (s)} \\ \midrule
\multicolumn{1}{c|}{1} & \multicolumn{1}{c|}{Feature} & \multicolumn{1}{c|}{-} & \multicolumn{1}{c|}{5222.35} & 1.26   \\
\multicolumn{1}{c|}{2} & \multicolumn{1}{c|}{Video} &\multicolumn{1}{c|}{ViCLIP} & \multicolumn{1}{c|}{16.42} & 422.87             \\
\multicolumn{1}{c|}{3} & \multicolumn{1}{c|}{Video} &\multicolumn{1}{c|}{CLIP}   & \multicolumn{1}{c|}{81.15} & 86.04             \\
\bottomrule[1pt]
\end{tabular}
\end{adjustbox}
\label{table6}
\end{table}

\noindent{\textbf{Analysis of Real-time Performance:}}
\textcolor{black}{
Table \ref{table6} evaluates the computational efficiency of VFEAL under both feature-input and end-to-end video-input settings, where latency measures the online localization process using pre-generated class-specific textual descriptions.
The LLM is invoked only once to generate these descriptions, and its outputs are cached and reused for all videos. Therefore, the LLM cost is a one-time preprocessing overhead rather than a repeated cost during streaming inference.
It is important to distinguish online causal inference from strict real-time processing.
The proposed framework is online in the sense that it only uses current and past frames, without requiring future information or retrospective refinement.
When pre-extracted visual features are used, VFEAL achieves efficient online localization, with a latency of only 1.26 seconds on videos with an average duration of 212.68 seconds.
However, under the end-to-end video-input setting, the ViCLIP-based implementation requires 422.87 seconds, and therefore does not satisfy strict real-time processing requirements.
The main bottleneck lies in visual feature extraction, as replacing ViCLIP with the lighter CLIP backbone reduces the latency to 86.04 seconds.
Therefore, the current framework supports causal online localization, while its strict real-time deployability depends on the efficiency of the adopted visual encoder.
}
\\

\begin{figure}
  \centering
  \includegraphics[width=\linewidth]{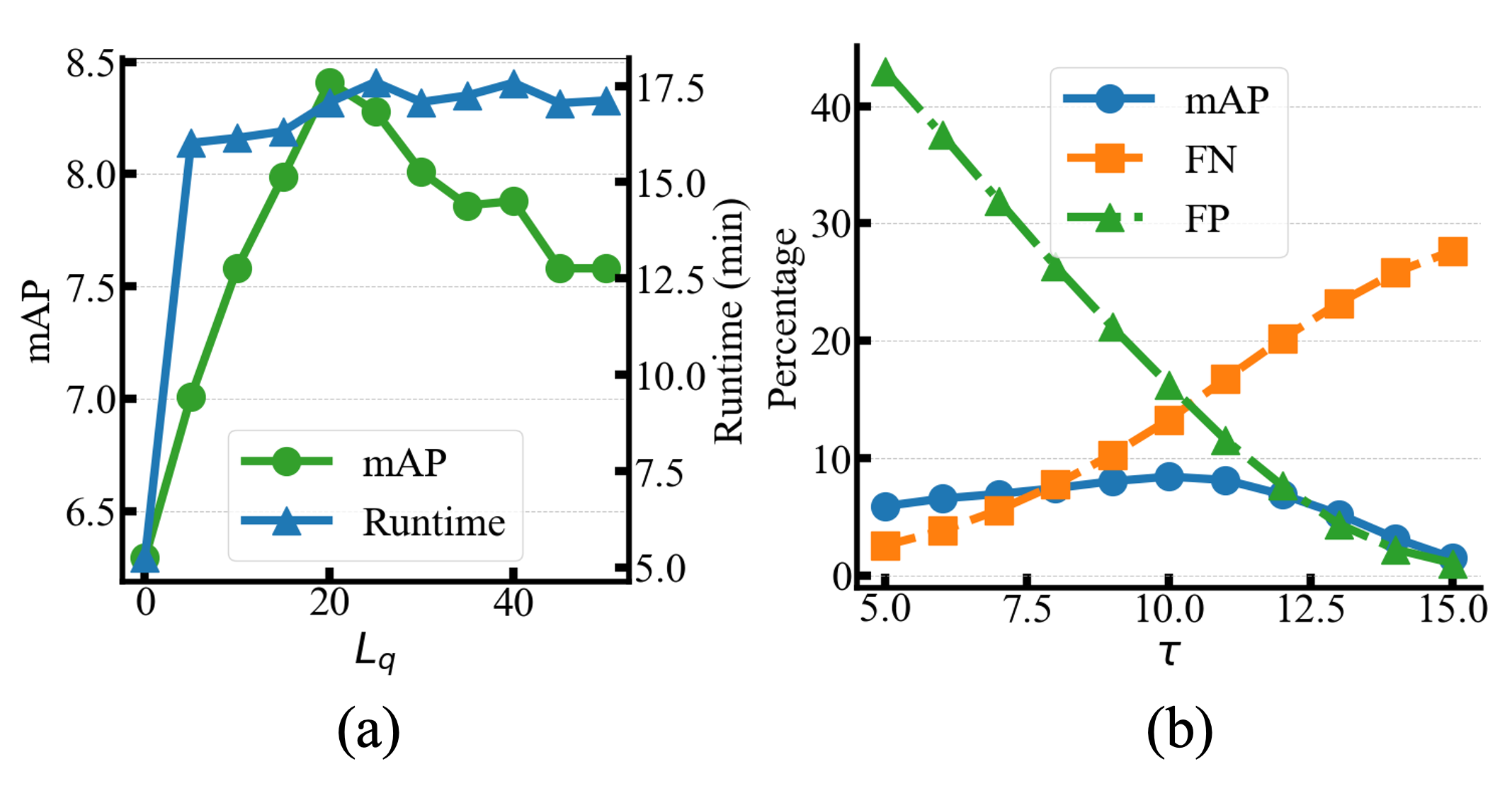}
  \caption{\textbf{Analysis of hyperparameters:} (a) memory bank length; (b) classification threshold.}
  \label{fig:ablation}
\end{figure}

\noindent{\textbf{Analysis of Memory Bank Length $L_q$:}}
We perform a sensitivity analysis on different memory bank lengths $L_q$ with a fixed similarity threshold of $\theta = 0.8$. As shown in Figure~\ref{fig:ablation}(a), incorporating the memory bank improves performance, with the best results achieved at a memory length of 20, which adequately covers most instances. This enhancement incurs an additional computation time of approximately 10 minutes for 24.4 hours of video, which is acceptable for streaming applications.\\

\noindent{\textbf{Analysis of Classification Threshold $\tau$:}}
The classification threshold $\tau$ plays a critical role in balancing over-detection and missed detections, directly impacting the quality of action span generation.
As shown in Figure \ref{fig:ablation}(b), mAP increases as $\tau$ ises from 5 to 10, reaching a peak of 8.41\% at $\tau=10$.
Increasing $\tau$ reduces the false positive rate, thereby decreasing the misclassification of non-action frames as actions. However, this also leads to a higher false negative rate, increasing the likelihood of missed action instances. Overall, the mAP metric demonstrates a degree of robustness with respect to the classification threshold.
\\

\noindent{\textbf{Analysis of Different Textual Setting:}}
As shown in Table~\ref{table4}, the fixed-text prompt yields the lowest performance, with an average mAP of 6.09\%, whereas using class-specific descriptions as prompts improves mAP by 0.5\%, indicating that class-specific prompts generated by LLMs provide more discriminative action features.
Row 3 corresponds to methods that treat the background as a separate class, which also performs worse, highlighting the importance of modeling background information for refining visual cues. Our background-aware approach improves mAP by 1.46\% by replacing the hard decision rule with a soft penalty, resulting in more reliable action span predictions.\\

\begin{table}[t]
\caption{Analysis of different classification strategies.}
\begin{adjustbox}{max width=\columnwidth}
\centering
\begin{tabular}{ccccccccc}
\toprule[1pt]
\textbf{Row} & \textbf{Text} & \multicolumn{1}{c|}{\textbf{BG}} & \textbf{0.3} & \textbf{0.4} & \textbf{0.5} & \textbf{0.6} & \multicolumn{1}{c|}{\textbf{0.7}} & \textbf{Avg} \\ \midrule
1            & fixed & \multicolumn{1}{c|}{\ding{55}}               & 12.96        & 8.49         & 5.13         & 2.64         & \multicolumn{1}{c|}{1.24}         & 6.09             \\
2            & class-specific  & \multicolumn{1}{c|}{\ding{55}} & 13.3         & 8.76         & 5.77         & 3.21         & \multicolumn{1}{c|}{1.89}         & 6.59             \\ \midrule
3            & $K+1$ & \multicolumn{1}{c|}{\ding{51}}                      & 14.05        & 9.43         & 5.96         & 3.44         & \multicolumn{1}{c|}{1.87}         & 6.95             \\
4            & BAKC & \multicolumn{1}{c|}{\ding{51}}                        & 17.69        & 11.6         & 7.04         & 3.88         & \multicolumn{1}{c|}{1.86}         & 8.41             \\ 
\bottomrule[1pt]
\end{tabular}
\end{adjustbox}
\label{table4}
\end{table}

\begin{figure}
  \centering
  \includegraphics[width=\linewidth]{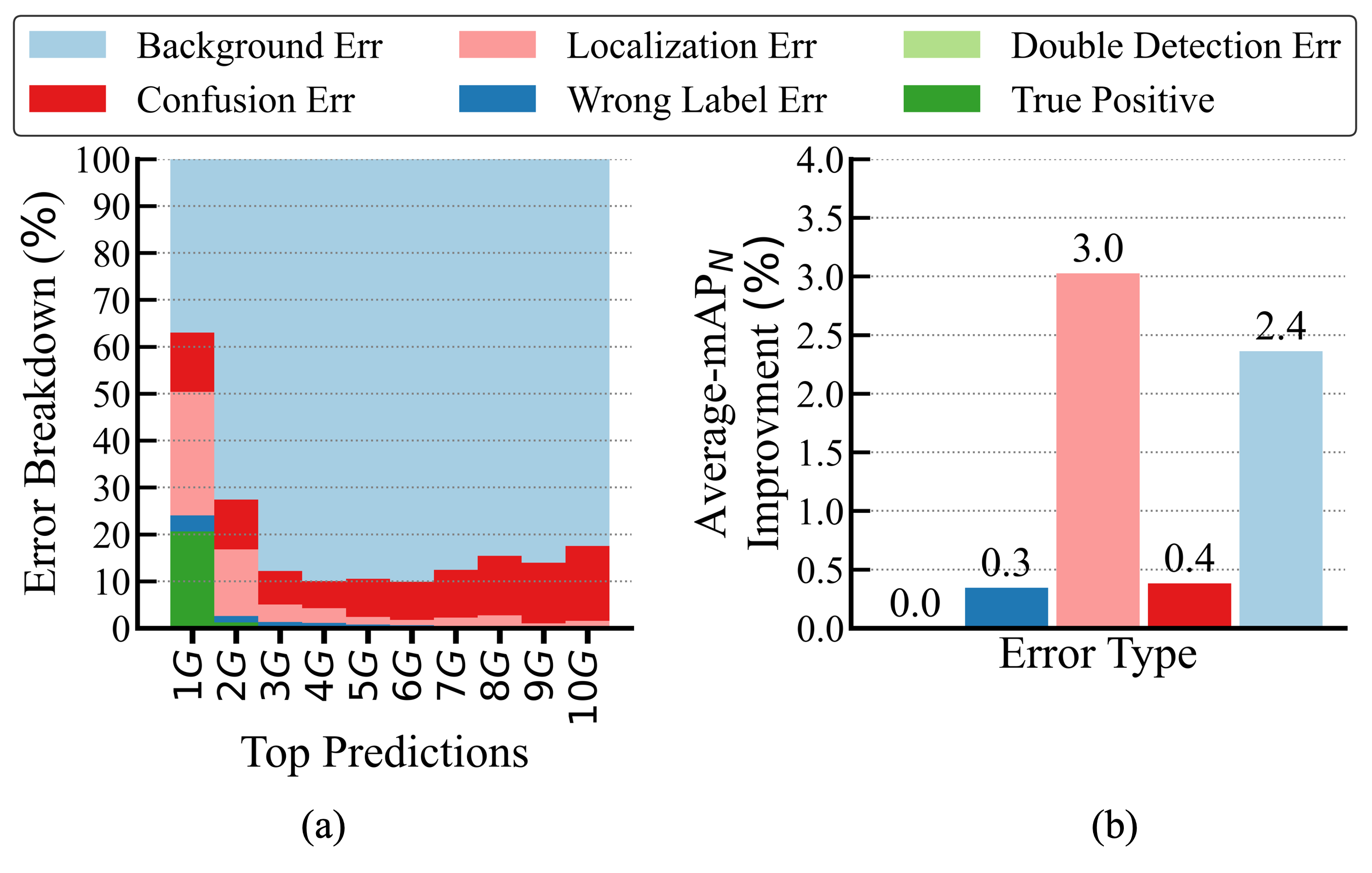}
  \caption{\textbf{Illustration of false positives:} (a) five sources of false positive errors, where $G$ denotes the total number of ground truth instances; and (b) their impact on average mAP improvement.}
  \label{fig:fp_analysis}
\end{figure}

\noindent{\textbf{Analysis of False Positives:}}
To assess the limitations of our model, a false positive analysis is conducted \cite{alwassel2018diagnosing} as illustrated in Figure~\ref{fig:fp_analysis}(a), where $G$ denotes the total number of ground truth instances. Background errors constitute the largest portion, suggesting a high degree of segment fragmentation. This issue is closely related to the threshold-based filtering mechanism. 
Figure~\ref{fig:fp_analysis}(b) shows that background and localization errors are the dominant sources of false positives, which should be addressed as key targets to further improve algorithm performance.\\

\begin{figure}[t]
  \centering
  \includegraphics[width=\linewidth]{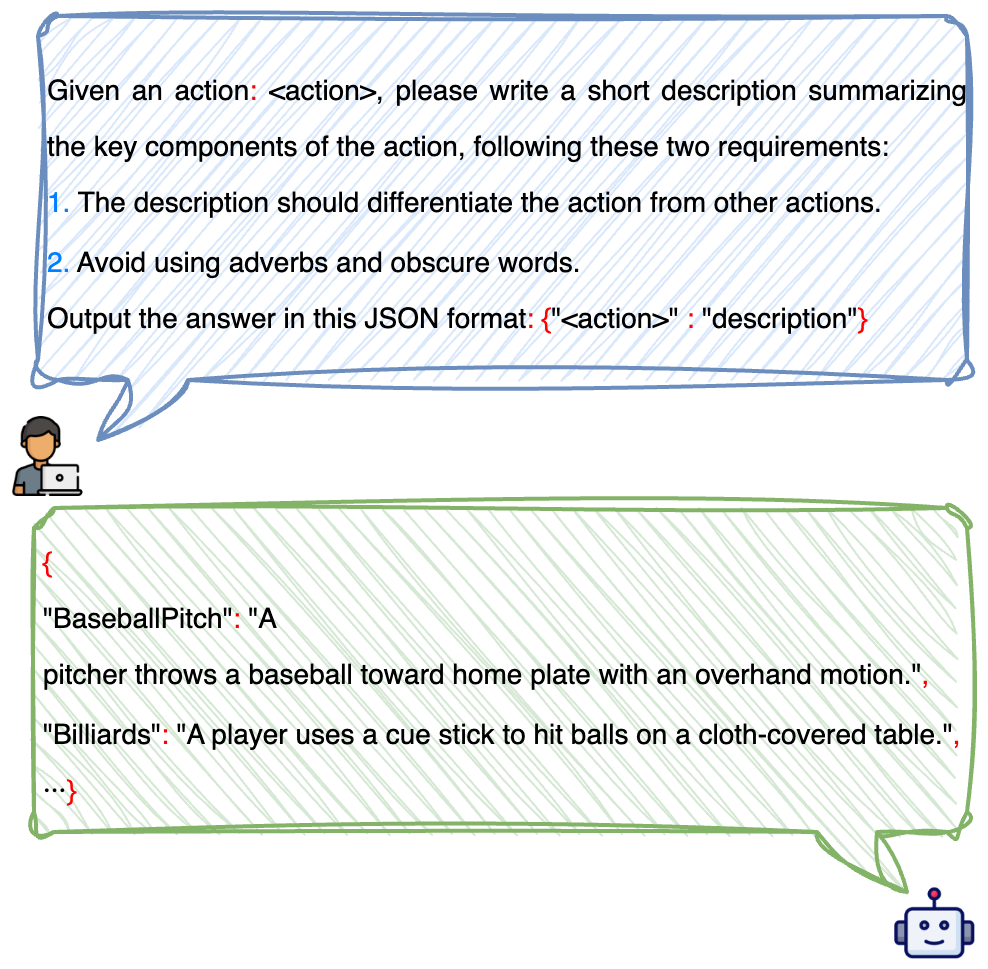}
  \caption{\textbf{Illustration of class-specific descriptions generated by an LLM.} The prompt provided to the LLM includes two generation requirements and one output format constraint.}
  \label{fig:prompt}
\end{figure}

\noindent{\textbf{Analysis of Different LLMs:}}
\label{section:prompt}
To facilitate zero-shot understanding \cite{zhang2020zstad,pourpanah2022review,wu2024towards,chen2025temporal}, we employ carefully designed prompt instructions with LLM to generate descriptive sentences for each action class, as illustrated in Figure~\ref{fig:prompt}. In this work, we impose two requirements: (1) to ensure distinction among action descriptions, improving feature discriminability, and (2) to avoid the use of adverbs and rare words, generating simple and concise descriptions. All generated action descriptions are stored in a dictionary format for convenient downstream use, where the keys correspond to action classes and the values are the generated textual descriptions.
Additionally, we evaluate different LLMs under the same prompt instructions, as shown in Table~\ref{table5}. Rows 2 and 3 correspond to GPT-4o \cite{hurst2024gpt} and DeepSeek-R1 \cite{guo2025deepseek}, respectively, using our refined prompts. The results indicate that the fixed description \textit{“this is a video of action [cls]”} performs poorly, and that different LLMs produce divergent results even when given identical prompts. 
While tailored prompts provide slight improvements over fixed descriptions, their benefit is limited on the THUMOS14 dataset due to its small number of coarse-grained action classes.
Nonetheless, class-specific prompts have demonstrated increasing effectiveness in the TAL community \cite{li2024detal,ju2022prompting,xu2025information} and remain a promising direction, particularly for datasets with finer-grained action categories. 
In our experiments, we adopt the descriptions generated by DeepSeek-R1, and with the support of our background-aware K-way classification, the final average mAP reaches 8.4\%.
\\

\begin{table}[t]
\caption{
Analysis of different LLMs.
}
\centering
\begin{adjustbox}{max width=\columnwidth}
\begin{tabular}{llcccccc}
\toprule[1pt]
\textbf{Row} & \textbf{Descriptions} & \textbf{0.3} & \textbf{0.4} & \textbf{0.5} & \textbf{0.6} & \textbf{0.7} & \textbf{Avg} \\ \midrule
1            & fixed          & 13.0         & 8.5          & 5.1          & 2.6          & 1.2          & 6.1          \\
2            & GPT-4o~\cite{hurst2024gpt}           & 13.8         & 9.1          & 6.0          & 3.2          & 2.1          & 6.8          \\
3            & DeepSeek-R1~\cite{guo2025deepseek}            & 13.3         & 8.8          & 5.8          & 3.2          & 1.9          & 6.6          \\ 
\bottomrule[1pt]
\end{tabular}
\end{adjustbox}
\label{table5}
\end{table}

\begin{figure*}[t]
  \centering
  \includegraphics[width=\linewidth]{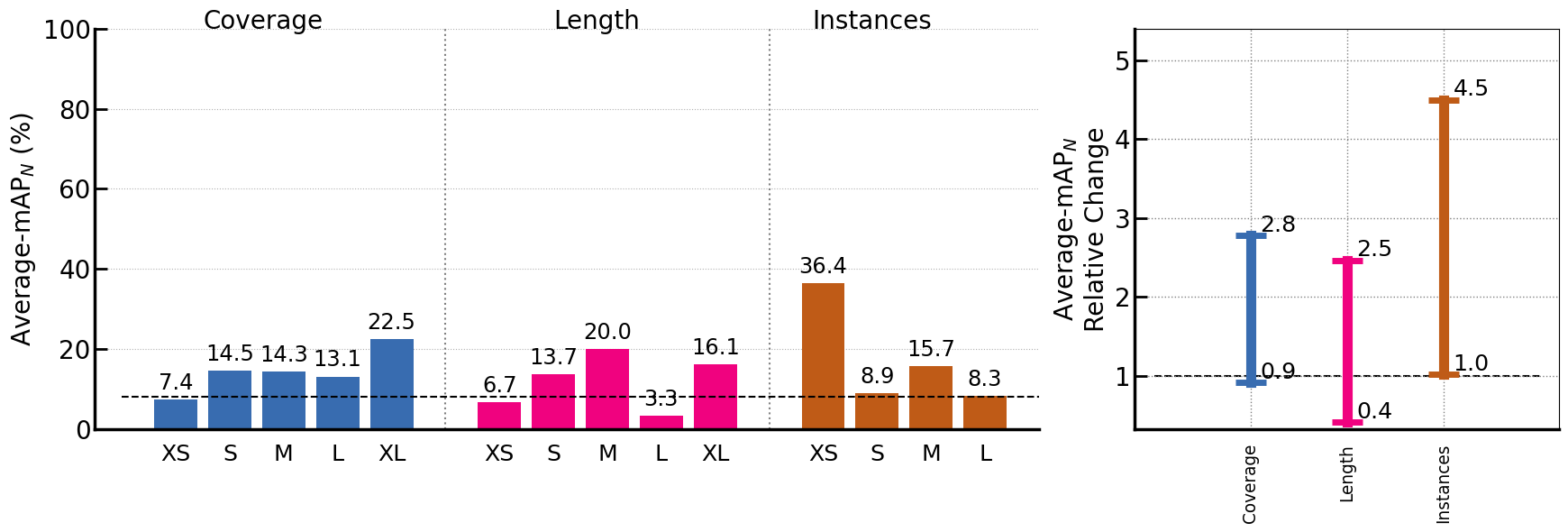}
  \caption{\textbf{Sensitivity analysis of VFEAL.} (a) Sensivity of VFEAL's average mAP to action characteristics. (b) The sensitivity profile summarizing the left figure. The difference between the max and min average-mAP\textsubscript{N} represents the sensitivity, while the difference between the max and the overall average-mAP\textsubscript{N} denotes the impact of the characteristic.}
\label{fig:sensitivity}
\end{figure*}

\begin{figure}[t]
  \centering
  \includegraphics[width=\linewidth]{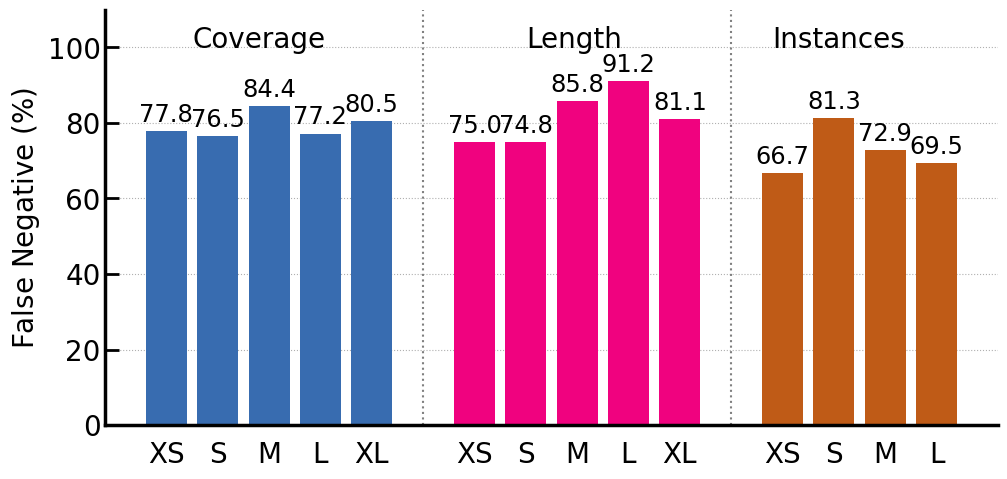}
  \caption{\textbf{False negative analysis.} Average false negative rate of VFEAL across three action characteristics on the THUMOS14 dataset.}
\label{fig:fn_analysis}
\end{figure}

\noindent{\textbf{Sensitivity Analysis:}}
Figure~\ref{fig:sensitivity}(a) presents a comprehensive sensitivity analysis of VFEAL on the THUMOS14 dataset, evaluating its performance across three action characteristics: coverage, length, and number of instances. The results are quantified using the normalized average mean Average Precision (mAP\textsubscript{N}) in percentage.
We begin by clearly defining the three action characteristics and their corresponding groups \cite{alwassel2018diagnosing}.
\textbf{Coverage} refers to the proportion of the video occupied by an action instance and is categorized into five bins: XS (0–20\%), S (20–40\%), M (40–60\%), L (60–80\%), and XL ($>$80\%). A higher coverage indicates that the action spans a larger fraction of the video.
\textbf{Length} denotes the absolute temporal duration (seconds) of an action instance, divided into: XS ($<$30s), S (30–60s), M (60–120s), L (120–180s), and XL ($>$180s).
\textbf{Instances} represents the number of same-class action occurrences within a single video, grouped as: XS (1 instance), S (2–4 instances), M (5–8 instances), and L ($>$8 instances).

The results reveal a clear and positive correlation between mAP\textsubscript{N} and coverage: model performance consistently improves as coverage increases, peaking at 22.5\% for XL actions ($>$80\% of the video duration) and dropping to 7.4\% for XS (0–20\%). This trend suggests that actions with broader temporal spans are inherently easier for the model to detect and accurately localize.
For action length, the best performance is observed in the M group (60–120s, 20.0\%), while a sharp and significant decline occurs for L (120–180s, 3.3\%). Extremely short actions (XS: 6.7\%) also result in notably poor performance, likely due to the lack of sufficient temporal cues. These results imply that both overly short and excessively long actions pose non-trivial challenges for accurate temporal localization.
Regarding the number of instances, the model achieves its highest performance on XS (single-instance videos, 36.4\%) and suffers a substantial decline as the instance count increases. In particular, performance deteriorates drastically in the L category ($>$8 instances, 8.3\%), highlighting the difficulty of distinguishing temporally overlapping actions in crowded video scenarios.
Figure~\ref{fig:sensitivity}(b) further reveals that VFEAL is more sensitive to the number of instances than to either coverage or length.\\

\noindent{\textbf{Analysis of False Negative:}}
We compute the proportion of missed detections for VFEAL and categorize the false negatives based on three action characteristics: coverage, length, and number of instances \cite{alwassel2018diagnosing}.
As illustrated in Figure~\ref{fig:fn_analysis}, VFEAL demonstrates robustness across different partitions of these characteristics.
Notably, the false negative rate reaches its lowest when the coverage falls within L (60–80\% of the video), the length is S (30–60s), and the number of instances is XS (1 instance). 
These findings indicate that actions with moderate coverage, short duration, and minimal instance overlap are more readily detected by VFEAL.
The results suggest that future work should prioritize improving performance on long-duration actions, densely populated scenes, and actions with extensive temporal coverage, in order to robustly handle varying action dynamics and complex visual contexts.\\
\begin{figure}[t]
  \centering
  \includegraphics[width=\linewidth]{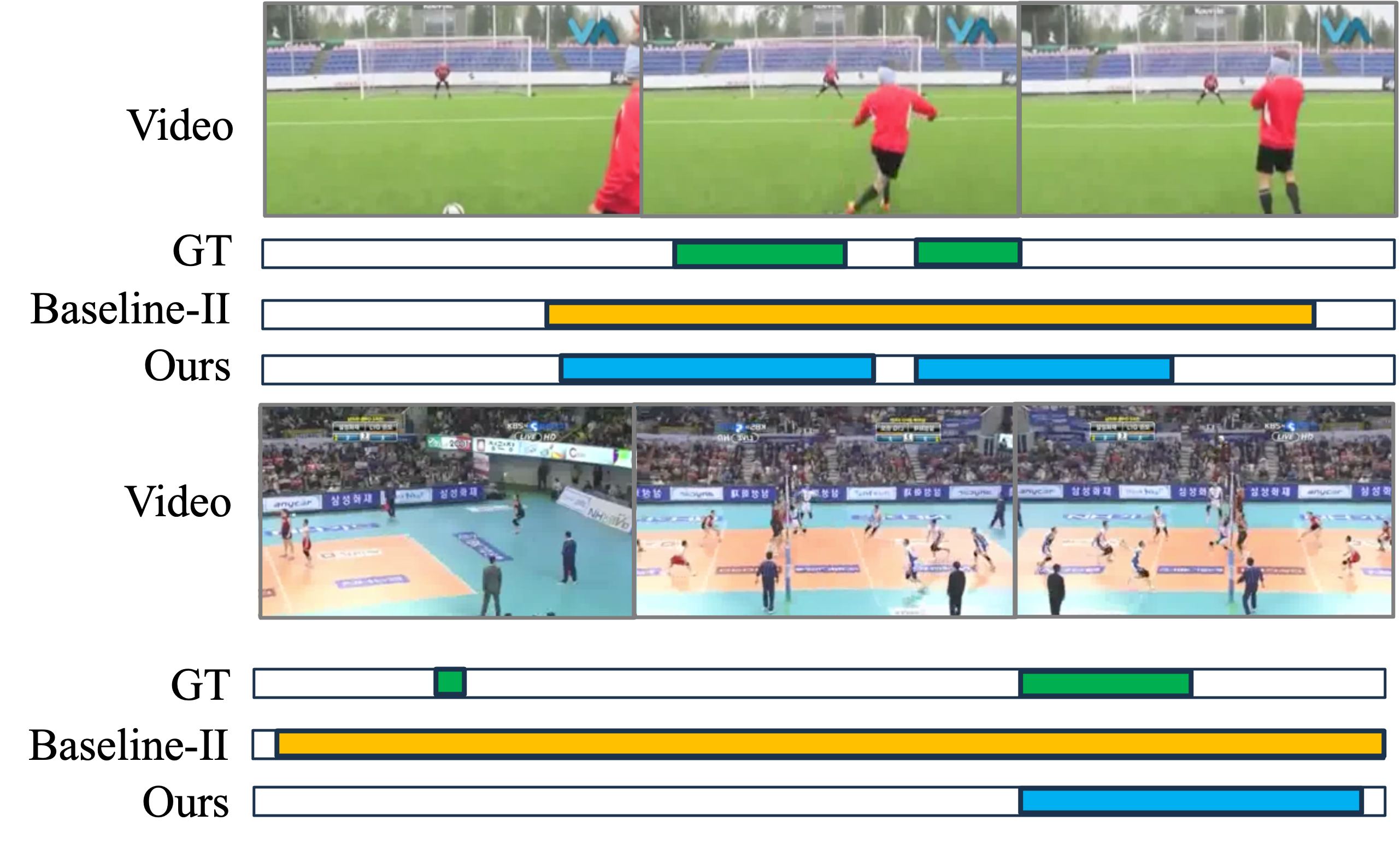}
  \caption{\textbf{Visualization of TAL Results.} Comparison of temporal localization outputs between Baseline-II and our method on two video examples from the THUMOS14 dataset.}
  \label{fig:visual}
\end{figure}

\noindent{\textbf{Visualization:}}
As illustrated in Figure~\ref{fig:visual}, we visualize the predictions of Baseline-II and our method for two annotated action instances, ``SoccerPenalty" and ``VolleyballSpiking", from the THUMOS14 dataset. The results show that Baseline-II struggles to accurately localize the temporal boundaries of these actions, whereas our method demonstrates a strong generalization capability to unseen action categories.\\



\section{Conclusion}
In this paper, we introduce a new task setting, online zero-shot temporal action localization, and establish new benchmarks on THUMOS14 and ActivityNet-1.3.
Our proposed action localizer consistently outperforms state-of-the-art offline and online methods.
Furthermore, we conduct extensive analyses to validate the effectiveness of the proposed memory-guided feature enhancement and background-aware k-way classification, along with in-depth examinations of false positives, false negatives, and model sensitivity.
We hope our work can inspire future research towards more real-time video understanding under open-world scenarios.


\small
\bibliographystyle{IEEEtran}
\bibliography{mybibfile-new}

\end{document}